\documentclass[conference]{IEEEtran}
\usepackage{amsmath}
\usepackage{amssymb}
\usepackage{tikz,pgfplots}
\usepackage{enumitem}
\usepackage{subfig}
\usepackage{booktabs}
\usepackage{algorithm, algorithmic}

\begin{document}

\title{Ensemble Kalman Filtering for Online Gaussian Process Regression and Learning}

\author{
	\IEEEauthorblockN{
		Danil Kuzin\IEEEauthorrefmark{1},
		Le Yang\IEEEauthorrefmark{2},
		Olga Isupova\IEEEauthorrefmark{3}
		and Lyudmila Mihaylova\IEEEauthorrefmark{1}
	}\\
	\IEEEauthorblockA{
		\IEEEauthorrefmark{1}Department of Automatic Control and Systems Engineering\\The University of Sheffield, Sheffield, UK\\Email: \{dkuzin1, l.s.mihaylova\}@sheffield.ac.uk
	}\\
	\IEEEauthorblockA{
		\IEEEauthorrefmark{2} Department of Electrical and Computer Engineering\\University of Canterbury, Christchurch, New Zealand\\E-mail: le.yang.le@gmail.com
	}\\
	\IEEEauthorblockA{
		\IEEEauthorrefmark{3}Department of Engineering Science\\The University of Oxford, Oxford, UK\\Email: olga.isupova@eng.ox.ac.uk
	}
}
\maketitle

\begin{abstract}
Gaussian process regression is a machine learning approach which has been shown its power for estimation of unknown functions. However, Gaussian processes suffer from high computational complexity, as in a basic form they scale cubically with the number of observations. Several approaches based on inducing points were proposed to handle this problem in a static context. These methods though face challenges with real-time tasks and when the data is received sequentially over time. In this paper, a novel online algorithm for training sparse Gaussian process models is presented. It treats the mean and hyperparameters of the Gaussian process as the state and parameters of the ensemble Kalman filter, respectively. The online evaluation of the parameters and the state is performed on new upcoming samples of data. This procedure iteratively improves the accuracy of parameter estimates. The ensemble Kalman filter reduces the computational complexity required to obtain predictions with Gaussian processes preserving the accuracy level of these predictions. The performance of the proposed method is demonstrated on the synthetic dataset and real large dataset of UK house prices.
\end{abstract}

\section{Introduction}
In Bayesian machine learning and signal processing, Gaussian processes (GPs) are used to approximate unknown functions~\cite{rasmussen2006gaussian} and provide posterior estimates for mean and variance of the target functions in the selected points. The function can be latent, and, in this case, GPs represent the idea of proximity, or a structure, when close values of inputs lead to close values of outputs. Another popular application is black-box optimisation with GPs, known as Bayesian optimisation. GPs are widely applied for signal processing, examples include audio~\cite{turner2011demodulation}, communications~\cite{perez2013gaussian}, and fault detection~\cite{svensson2015marginalizing}.

GPs are characterised by covariance functions that usually have a set of hyperparameters. The popular examples are squared-exponential, Mat\'ern and exponential covariance functions~\cite{rasmussen2006gaussian}. They are stationary functions that depend only on distance between points, they provide solutions with different smoothness properties. The hyperparameters are hard to estimate by experts and they are usually learnt within the GP framework, for example by optimising the marginal likelihood, which leads to local maxima.

GPs are usually represented in a grid of points and it is the source of the main limitation. The required resources are huge: computational time scales cubically with the number of grid points, required memory scales quadratically. It is essential to reduce these numbers in order to make GPs applicable for larger datasets or online inference.

During the last decades multiple approaches have been proposed to deal with this problem. The most popular approach is introduction of inducing points~\cite{quinonero2005unifying} where the locations of grid points are optimised, their amount is reduced with an attempt to maintain good predictive power. In~\cite{titsias2009variational} inducing points are treated as variational parameters and Bayesian inference is performed. In~\cite{bui2016unifying} expectation propagation is proposed for the Bayesian inference of the parameters.

Another approach is the distributed computations, where local predictions are combined into unified mean and variance predictions. The idea of partitioning dataset for the GP problem is considered in~\cite{shen2006fast} with use of Kd-trees. The distributed Bayesian version with sparse approximation is proposed in~\cite{gal2014distributed}.

In~\cite{huber2014recursive} the online procedure for updating GP parameters is proposed. The mean in the grid points is treated as a state variable, GP hyperparameters and noise are treated as parameters and for the joint state-parameter vector the unscented Kalman filter is used. The model has been recently used for the received-signal-strength estimation~\cite{yin2017received, yin2017distributed}, flow modelling and prediction in sports analytics~\cite{zhao2016gaussian}. Sampling approaches for the online updating of the GP hyperparameters include slice sampler~\cite{murray2010slice}, sequential Monte Carlo~\cite{svensson2015marginalizing}, Bayesian Monte Carlo~\cite{osborne2012real}.

In this paper the ensemble Kalman filter is proposed to deal with online GP problem. It provides more stable parameter estimates with better predictive performance.

The main contributions of this paper can be summarised as:
\begin{itemize}
\item For the first time the ensemble Kalman filter (EnKF) for the problem of online GP regression and learning is proposed. This allows to reduce the computational complexity related to the prediction, as the size of the invertible matrices is reduced according to the ensemble sizes.
\item The dual and joint versions of the ensemble Kalman filter are presented in the paper.
\item The performance of the algorithms is compared using the synthetic dataset and real large dataset of the house prices.
\end{itemize}

The paper is organised in the following way: first the overview of the ensemble Kalman filter and the problem of state and parameter estimation within this framework is described in Section~\ref{sec:enkf_overview}. In Section~\ref{sec:dual_enkf_gp} the proposed joint and dual EnKF frameworks for GPs are described. In Section~\ref{sec:experiments} the experiments are conducted on the synthetic and UK house price data, the conclusions are presented in Section~\ref{sec:conclusion}.

\section{Ensemble Kalman filter overview}
\label{sec:enkf_overview}
Ensemble Kalman filter was originally discussed in~\cite{evensen1994sequential}, and a recent overview with different improvement techniques is given in~\cite{roth2017ensemble}. EnKF uses the Monte Carlo method to generate an ensemble of state sigma points and then this state ensemble is passed through the measurement function to obtain the observation ensemble, it is additionally perturbed with the measurement noise. The mean and variance of the resulting observational distribution together with actual observations are used to update the state. The main computational difference in comparison to the classic Kalman filter is that the covariance matrices are replaced with ensembles that can be less in dimensionality.

The usual approach to parameters estimation is augmenting the state vector with parameters vector thus creating the larger augmented state-parameter vector. It can then be used to perform the online estimation within the EnKF framework~\cite{anderson2001ensemble, evensen2009ensemble}. In~\cite{wan1997dual} dual estimation of state and parameters is proposed to replace joint estimation as in classic Kalman filters: for every new observation, first the parameters are updated and then using the updated parameters the state is updated. Dual estimation of parameters and state for EnKF is considered in~\cite{moradkhani2005dual}.

Other approaches for parameters estimation in EnKF include the maximum likelihood method~\cite{mitchell2000adaptive, delsole2010state} and the Bayesian inference~\cite{stroud2007sequential}.

\section{Ensemble Kalman Filter for Gaussian Processes}
\label{sec:dual_enkf_gp}
This paper proposes the algorithm for the problem of online estimation of the constant unknown continuous function $f(\mathbf{x})$ of the $D$-dimensional input vector $\mathbf{x} \in \mathbb{R}^D$. The unknown function is approximated with a GP: the mean $\mathbf{g}\in\mathbb{R}^K$ of the GP is approximated at the $K$ grid points $\mathbf{X}_g\in\mathbb{R}^{K \times D}$ and $L_\theta$ parameters of the covariance function $\boldsymbol\theta \in \mathbb{R}^{L_\theta}$ are estimated. With the mean and parameters of the covariance function it is possible to predict the mean and variance of $f(\mathbf{x^*})$ at any point $\mathbf{x^*}$.

It is assumed that the observations of the function are available sequentially, at every timestamp $1 \le t \le T$, where $T$ is the last observation timestamp. At every iteration, $t$ of the algorithm a total of $S$ one-dimensional noisy function observations $\mathbf{y}_t \in \mathbb{R}^S$ are obtained at random points $\mathbf{x}_\text{new}$. The variance of noise $\sigma^2_y$ assumed to be unknown and it is estimated at every iteration of the algorithm. The full vector of parameters is therefore $\boldsymbol\eta = [\boldsymbol\theta, \sigma^2_y] \in \mathbb{R}^L$, where $L = L_\theta + 1$.

The dependency between covariance function parameters and observations is non-linear, therefore a nonlinear version of Kalman filter is required. The ensemble Kalman filter allows to have constant complexity for updates, which is determined by the number of ensemble points, $N$.

Two versions of ensemble Kalman filter for the online GP learning are proposed, they differ in the way how hyperparameters of the GP are treated: \textit{Dual EnKF} first updates the hyperparameters of the GP and then based on their estimates updates the state; \textit{Joint EnKF} updates hyperparameters of the GP and the state simultaneously with the augmented state--hyperparameter vector.

\subsection{Dual Ensemble Kalman Filter for Gaussian Processes}
\label{sec:dual_dual_enkf_gp}
This algorithm is further denoted as Dual GP-EnKF. It uses the ensembles of same size $N$ to approximate the distributions of the parameters and state. At every iteration the predicted distributions of the parameters and state are computed, and the observations are predicted. Then, based on the cross-covariance of the parameter and observation ensembles, Kalman gain is computed and it is used to update the parameter distribution. After this step, new observations are predicted with updated parameters and then the cross-covariance of new observations and the state is used to update the state. The details of the algorithm are presented below.
\subsubsection{Initialisation}
Initially, ensembles for the parameters $ \mathbf{H} \in \mathbb{R}^{N \times L} = [\boldsymbol\eta^{(i)} \in \mathbb{R}^{1 \times L}]_{1 \le i\le N} $ and mean $ \mathbf{G}\in \mathbb{R}^{N \times K} = [\mathbf{g}^{(i)} \in \mathbb{R}^{1\times K}]_{1 \le i\le N} $ at the grid points of the GP are generated. The rows of matrices correspond to the ensemble members. For parameters that can only be positive, such as the variance, logarithms of their values are used in the ensemble. Initial ensembles are generated from the Gaussian distribution: for each ensemble index $1 \le i\le N$
\begin{subequations}
\label{eq:init}
\begin{align}
	\boldsymbol\eta^{(i)}_{0 | 0} & \sim \mathcal{N}(\mathbf{0}, \boldsymbol\Sigma_H), \\
	\mathbf{g}^{(i)}_{0 | 0} & \sim \mathcal{N}(\mathbf{0}, \boldsymbol\Sigma_G),
\end{align}
\end{subequations}
where $\boldsymbol\Sigma_H$, $\boldsymbol\Sigma_G$ are the initial covariance matrices for the ensembles. In our experiments, they are assumed to be diagonal. $\mathcal{N}(\cdot)$ denotes the Gaussian distribution.

After the initialisation at every iteration of the algorithm three steps follow: prediction, update for parameters, update for state.

\subsubsection{Prediction}
For the whole running time of the algorithm the estimated function remains constant, while unknown. This can be simulated with the random walk motion model for the parameters and state, each ensemble member is updated as
\begin{subequations}
\label{eq:predict_params_state}
\begin{align}
	\boldsymbol\eta^{(i)}_{t+1 | t}  &= \boldsymbol\eta^{(i)}_{t | t} + \boldsymbol\varepsilon_\eta, \\
	\mathbf{g}^{(i)}_{t+1 | t} &= \mathbf{g}^{(i)}_{t | t} + \boldsymbol\varepsilon_g,
\end{align}
\end{subequations}
where $\boldsymbol\varepsilon_\eta \sim \mathcal{N}(\mathbf{0}, \sigma_\eta\mathbf{I})$, $\boldsymbol\varepsilon_g \sim \mathcal{N}(\mathbf{0}, \sigma_g\mathbf{I})$ are the noise variables with corresponding variances. We also consider Liu-West filter that has other approach to parameter prediction later in this section.

Assume that $S$ observations were obtained at locations $\mathbf{X}_\text{new} = [\mathbf{x}^s_\text{new}]_{s=1}^S$. According to the definition of GPs the joint distribution for any discrete set of samples is the multivariate Gaussian distribution. Therefore, for each parameter ensemble $i$ the distribution of predicted function values $\widehat{\mathbf{y}}^{(i)} = [\widehat{y}^{1, (i)}, \ldots, \widehat{y}^{S, (i)}]$ at locations $\mathbf{x}_\text{new}$ can be obtained as
\begin{align}
\label{eq:predict_obs}
\begin{split}
\widehat{\mathbf{y}}^{(i)} = {}& K(\mathbf{X}_\text{new}, \mathbf{X}_g | \boldsymbol\theta^{(i)}_{t+1 | t} )\\
				    &\times[K(\mathbf{X}_g, \mathbf{X}_g| \boldsymbol\theta^{(i)}_{t+1 | t}) + \sigma^{2 (i)}_{y\,t+1 | t}\mathbf{I}]^{-1}\mathbf{g}^{(i)}_{t+1 | t},
\end{split}
\end{align}
where $K(\mathbf{X}_1, \mathbf{X}_2 | \boldsymbol\theta)$ is the covariance matrix evaluated at every pair of points from $\mathbf{X}_1$, $\mathbf{X}_2$ with parameters~$\boldsymbol\theta$; $\boldsymbol\theta^{(i)}_{t+1 | t}$ and $\sigma^{2 (i)}_{y\,t+1 | t}$ are components of the joint parameter vector $\boldsymbol\eta^{(i)}_{t+1 | t}$. The matrix for all predictions is denoted as $\widehat{\mathbf{Y}} \in \mathbb{R}^{N \times S} = [\widehat{\mathbf{y}}^{(i)}]_{i=1}^{N}$

In EnKF, observations are treated as random variables and the observation ensemble is generated, which has a Gaussian distribution around the actual observation with predefined covariance $\sigma^2_{\text{obs}}$
\begin{equation}
\label{eq:compute_traj}
	\mathbf{y}^{(i)} = \mathbf{y} + \boldsymbol\varepsilon_{\text{obs}},
\end{equation}
where $\boldsymbol\varepsilon_{\text{obs}} \sim \mathcal{N}(0, \sigma^2_{\text{obs}}\mathbf{I})$

\subsubsection{Update parameters}
EnKF updates are similar to the usual KF, with the means and covariances estimated from the ensembles.
First, cross covariances of the parameter ensemble and prediction ensemble are computed. Let $\mathbb{E}_i[\cdot]$ denote the expected value with respect to ensembles. Then
\begin{subequations}
\label{eq:cross_cov_param}
\begin{align}
	\overline{\boldsymbol{\eta}}_{t+1 | t} = &\frac{1}{N}\sum_{i=1}^N\boldsymbol\eta^{(i)}_{t+1 | t},\\
	\begin{split}
	\boldsymbol\Sigma^{\boldsymbol\eta \mathbf{y}} = &\mathbb{E}_i\left[(\mathbf{H}_{t+1 | t} - \mathbb{E}_i[\mathbf{H}_{t+1 | t}])^\top(\widehat{\mathbf{Y}} - \mathbb{E}_i[\widehat{\mathbf{Y}}])\right] \\
	= &\frac{1}{N-1}\sum_{i=1}^N(\boldsymbol\eta^{(i)}_{t+1 | t} - \overline{\boldsymbol\eta}_{t+1 | t})^\top(\widehat{\mathbf{y}}^{(i)} - \mathbf{y})
	\end{split}
\end{align}
\end{subequations}

After that, the covariance matrix of the predictions is computed
\begin{align}
\label{eq:cross_cov_pred}
\begin{split}
	\boldsymbol\Sigma^{\mathbf{y} \mathbf{y}} = &\mathbb{E}_i\left[(\widehat{\mathbf{Y}} - \mathbb{E}_i[\widehat{\mathbf{Y}}])^\top(\widehat{\mathbf{Y}} - \mathbb{E}_i[\widehat{\mathbf{Y}}])\right] \\
	= & \frac{1}{N-1}\sum_{i=1}^N(\widehat{\mathbf{y}}^{(i)} - \mathbf{y})^\top(\widehat{\mathbf{y}}^{(i)} - \mathbf{y})
\end{split}
\end{align}

Then the Kalman gain for correcting the parameters can be computed as
\begin{equation}
\label{eq:kalman_param}
	\mathbf{K}^{\boldsymbol\eta} = \boldsymbol\Sigma^{\boldsymbol\eta \mathbf{y}}(\boldsymbol\Sigma^{\mathbf{y} \mathbf{y}} +\sigma^2_{\text{obs}} \mathbf{I})^{-1}
\end{equation}
The parameters are updated as
\begin{equation}
\label{eq:update_param}
	\boldsymbol\eta^{(i)}_{t+1 | t+1} = \boldsymbol\eta^{(i)}_{t+1 | t} + \mathbf{K}^{\boldsymbol\eta}(\mathbf{y}^{(i)} - \widehat{\mathbf{y}}^{(i)})
\end{equation}

\subsubsection{Update state}
Updates for the state are similar to the updates for parameters, but with the updated values of parameters.

First, predictions of observations are corrected with updated parameters using
\begin{align}
\label{eq:predict_obs2}
\begin{split}
\widehat{\mathbf{y}}^{(i)} = &{} K(\mathbf{X}_\text{new}, \mathbf{X}_g | \boldsymbol\theta^{(i)}_{t+1 | t+1} )\\
					  &\times [K(\mathbf{X}_g, \mathbf{X}_g| \boldsymbol\theta^{(i)}_{t+1 | t+1}) + \sigma^{2 (i)}_{y\,t+1 | t+1}\mathbf{I}]^{-1}\mathbf{g}^{(i)}_{t+1 | t}
\end{split}
\end{align}

After that, the cross covariance of the state ensemble and prediction ensemble is updated
\begin{subequations}
\label{eq:cross_cov_state}
\begin{align}
	\overline{\mathbf{g}}_{t+1 | t} = &\frac{1}{N}\sum_{i=1}^N\mathbf{g}^{(i)}_{t+1 | t},\\
	\begin{split}
		\mathbf{\Sigma}^{\mathbf{g} \mathbf{y}} = &\mathbb{E}_i\left[(\mathbf{G}_{t+1 | t} - \mathbb{E}[\mathbf{G}_{t+1 | t}])^\top(\widehat{\mathbf{Y}} - \mathbb{E}[\widehat{\mathbf{Y}}])\right] \\
		= &\frac{1}{N-1}\sum_{i=1}^N(\mathbf{g}^{(i)}_{t+1 | t} - \overline{\mathbf{g}}_{t+1 | t})^\top(\widehat{\mathbf{y}}^{(i)} - \mathbf{y})
	\end{split}
\end{align}
\end{subequations}

After that, the covariance matrix of the predictions is computed according to~(\ref{eq:cross_cov_pred}) and then the Kalman gain for correcting the state
\begin{equation}
\label{eq:kalman_state}
	\mathbf{K}^\mathbf{g} = \boldsymbol\Sigma^{\mathbf{g} \mathbf{y}}(\boldsymbol\Sigma^{\mathbf{y} \mathbf{y}} + \sigma^2_{\text{obs}} \mathbf{I})^{-1}
\end{equation}

Then the state is updated as
\begin{equation}
\label{eq:update_state}
	\mathbf{g}^{(i)}_{t+1 | t+1} = \mathbf{g}^{(i)}_{t+1 | t} + \mathbf{K}^{\mathbf{g}}(\mathbf{y}^{(i)} - \widehat{\mathbf{y}}^{(i)})
\end{equation}

The resulting algorithm is given in~\figurename\ref{fig:algorithm}.

\begin{figure}[t]
  \centering
    \fbox{
  	\parbox{0.95\columnwidth}{
	\begin{enumerate}[leftmargin=12pt]
	\item Initialise (\ref{eq:init})
	\item Iterate
    		\begin{enumerate}[leftmargin=8pt]
			\def\labelenumi{\arabic{enumi}.}
			\item Predict, for each $1 \le i\le N$:
			\begin{itemize}[leftmargin=8pt]
  				\item[P1] Predict the parameters and state (\ref{eq:predict_params_state})
  				\item[P2] Predict observations (\ref{eq:predict_obs})
  				\item[P3] Compute noisy trajectories (\ref{eq:compute_traj})
  			\end{itemize}
			\item Update parameters
			\begin{itemize}
  				\item[UP1] Compute cross covariance of the parameter ensemble and prediction ensemble (\ref{eq:cross_cov_param})
  				\item[UP2] Compute covariance matrix of the predictions (\ref{eq:cross_cov_pred})
  				\item[UP3] Compute Kalman gain for correcting the parameters (\ref{eq:kalman_param})
  				\item[UP4] Update the parameters (\ref{eq:update_param})
  			\end{itemize}
 			\item Update model state
 			\begin{itemize}
 				\item[US1] Predict observations with updated parameters (\ref{eq:predict_obs2})
  				\item[US2] Compute cross covariance of the state ensemble and prediction ensemble (\ref{eq:cross_cov_state})
				\item[US3] Compute covariance matrix of the predictions (\ref{eq:cross_cov_pred})
				\item[US4] Compute Kalman gain for correcting the state (\ref{eq:kalman_state})
  				\item[US5] Update the state (\ref{eq:update_state})
  			\end{itemize}
		\end{enumerate}
	\end{enumerate}
  	}
	}
  \caption{Dual GP-EnKF algorithm}
  \label{fig:algorithm}
\end{figure}

\subsection{Liu-West filter}
\label{sec:lw_enkf_gp}
The evolution of the parameter distribution in (\ref{eq:predict_params_state}) leads to its over-diffuse. The Liu-West filter~\cite{liu2001combined} uses kernel density estimation, it can be used to estimate the predicted distribution of parameters so that the resulting distribution converges to the true distribution. It is parametrised with the discount factor $\delta_{lw} \in (0, 1]$, that is usually taken from the interval $[0.95, 0.99]$. With introduction of additional parameters
\begin{subequations}
\begin{align}
a_{lw} &= \frac{3 \delta_{lw} - 1}{2 \delta_{lw}}, \\
h_{lw}^2 &= 1 - a_{lw}^2,
\end{align}
\end{subequations}

the evolution of the parameter density is
\begin{equation}
\boldsymbol\eta^{(i)}_{t+1 | t}  = a_{lw}\boldsymbol\eta^{(i)}_{t | t} + (1 - a_{lw})\overline{\mathbf{\eta}}_{t | t} + \boldsymbol\varepsilon_{lw},
\end{equation}
where $\boldsymbol\varepsilon_{lw} \sim \mathcal{N}(\mathbf{0}, \sqrt{h_{lw}^2\operatorname{Var}\boldsymbol\eta_{t | t}})$
The algorithm is further denoted as Liu-West Dual GP-EnKF.

\subsection{Joint Ensemble Kalman Filter for Gaussian Processes}
\label{sec:joint_enkf_gp}
It is also possible to estimate parameters of the model by augmenting the state vector $\mathbf{g}$ by the parameters vector $\boldsymbol\eta$: the augmented state is $\boldsymbol{s} = [ \mathbf{g}; \boldsymbol\eta]$. The algorithm is further denoted as Joint GP-EnKF. The details are presented in \figurename\ref{pic:augmented_algorithm}.

\subsubsection{Initialisation}
Initially, an ensemble for the augmented state $ \mathbf{S} \in \mathbb{R}^{N \times (L+K)} = [\mathbf{s}^{(i)}]_{1 \le i\le N} $ is generated. For each $1 \le i\le N$
\begin{subequations}
\label{eq:init_augmented}
\begin{align}
	\boldsymbol\eta^{(i)}_{0 | 0} & \sim \mathcal{N}(\mathbf{0}, \boldsymbol\Sigma_S)
\end{align}
\end{subequations}
where $\boldsymbol\Sigma_S$ is the initial covariance matrices for the ensembles.
After the initialisation the algorithm iterates the prediction and update steps.

\subsubsection{Prediction}
Similar to the dual EnKF, the random walk assumption for the motion model of the augmented state is used. Each ensemble member is updated as
\begin{equation}
\label{eq:predict_params_state_augmented}
	\mathbf{s}^{(i)}_{t+1 | t} = \mathbf{s}^{(i)}_{t | t} + \boldsymbol\varepsilon_s,
\end{equation}
where $\boldsymbol\varepsilon_s \sim \mathcal{N}(\mathbf{0}, \sigma_s\mathbf{I})$ is the noise variable with corresponding variance.

The predictions are made in a same way as in (\ref{eq:predict_obs}) and observations are noised as in (\ref{eq:compute_traj}).

\subsubsection{Update}
Updates for the augmented state are similar to the updates for the state in dual EnKF. The cross covariance of the augmented state ensemble and prediction ensemble is estimated as
\begin{subequations}
\label{eq:cross_cov_state_augmented}
\begin{align}
	\overline{\mathbf{s}}_{t+1 | t} = &\frac{1}{N}\sum_{i=1}^N\mathbf{s}^{(i)}_{t+1 | t},\\
	\begin{split}
		\mathbf{\Sigma}^{\mathbf{s} \mathbf{y}} = &\mathbb{E}_i\left[(\mathbf{S}_{t+1 | t} - \mathbb{E}[\mathbf{S}]_{t+1 | t})^\top(\widehat{\mathbf{Y}} - \mathbb{E}[\widehat{\mathbf{Y}}])\right] \\
		= &\frac{1}{N-1}\sum_{i=1}^N(\mathbf{s}^{(i)}_{t+1 | t} - \overline{\mathbf{s}}_{t+1 | t})^\top(\widehat{\mathbf{y}}^{(i)} - \mathbf{y})
	\end{split}
\end{align}
\end{subequations}

After that, the covariance matrix of the predictions is computed as (\ref{eq:cross_cov_pred}) and then the Kalman gain for correcting the augmented state
\begin{equation}
\label{eq:kalman_state_augmented}
	\mathbf{K}^\mathbf{s} = \boldsymbol\Sigma^{\mathbf{s} \mathbf{y}}(\boldsymbol\Sigma^{\mathbf{y} \mathbf{y}} + \sigma_y \mathbf{I})^{-1}
\end{equation}

Then the augmented state is updated as
\begin{equation}
\label{eq:update_state_augmented}
	\mathbf{s}^{(i)}_{t+1 | t+1} = \mathbf{s}^{(i)}_{t+1 | t} + \mathbf{K}^{\mathbf{s}}(\mathbf{y}^{(i)} - \widehat{\mathbf{y}}^{(i)})
\end{equation}

\begin{figure}[t]
  \centering
    \fbox{
  	\parbox{0.95\columnwidth}{
	\begin{enumerate}[leftmargin=12pt]
	\item Initialise (\ref{eq:init_augmented})
	\item Iterate
    		\begin{enumerate}[leftmargin=8pt]
			\def\labelenumi{\arabic{enumi}.}
			\item Predict, for each $1 \le i\le N$:
			\begin{itemize}[leftmargin=8pt]
  				\item[P1] Predict the augmented state (\ref{eq:predict_params_state_augmented})
  				\item[P2] Predict observations (\ref{eq:predict_obs})
  				\item[P3] Compute noisy trajectories (\ref{eq:compute_traj})
  			\end{itemize}
 			\item Update the augmented state
 			\begin{itemize}
  				\item[US1] Compute cross covariance of the augmented state ensemble and the prediction ensemble (\ref{eq:cross_cov_state_augmented})
				\item[US2] Compute covariance matrix of the predictions (\ref{eq:cross_cov_pred})
				\item[US3] Compute Kalman gain for correcting the augmented state (\ref{eq:kalman_state_augmented})
  				\item[US4] Update the augmented state (\ref{eq:update_state_augmented})
  			\end{itemize}
		\end{enumerate}
	\end{enumerate}
  	}
	}
  \caption{Joint GP-EnKF algorithm}
  \label{pic:augmented_algorithm}
\end{figure}

\subsection{Computational complexity of Kalman Filter approaches for Gaussian Processes}
\paragraph{Dual Ensemble Kalman Filter}
At the prediction step the most demanding operation is the prediction of observations, that requires inversion of the covariance matrix for each ensemble member, that is $\mathcal{O}(NK^3)$. At the update steps it is computation of Kalman gains, that is $\mathcal{O}(S^3) + \mathcal{O}(LS^2)$ for the parameters and $\mathcal{O}(S^3) + \mathcal{O}(KS^2)$ for the state.
The resulting computational time complexity for the  Dual GP-EnKF is $\mathcal{O}(T (NK^3 + S^3 + (L + K)S^2))$

\paragraph{Joint Ensemble Kalman Filter}
Joint EnKF has the same asymptotic complexity as Dual EnKF.

\paragraph{Classic GP}
The classic GP without inducing points that stores all previous observations and recomputes predictions at every time step $t$ has $\mathcal{O}(S^3t^3)$ computational complexity due to the covariance matrix growing in size as $St$ at every dimension. Note that the computational complexity of all GP-EnKF algorithms is linear with respect to the number~$T$ of time steps.

\section{Experiments}
\label{sec:experiments}

\begin{figure}[t]
\centering
	\subfloat[The target function used in the synthetic data experiment. As an example, the sample from the function used as the input for the algorithms at one iteration is displayed.]{\includegraphics[width=0.45\textwidth]{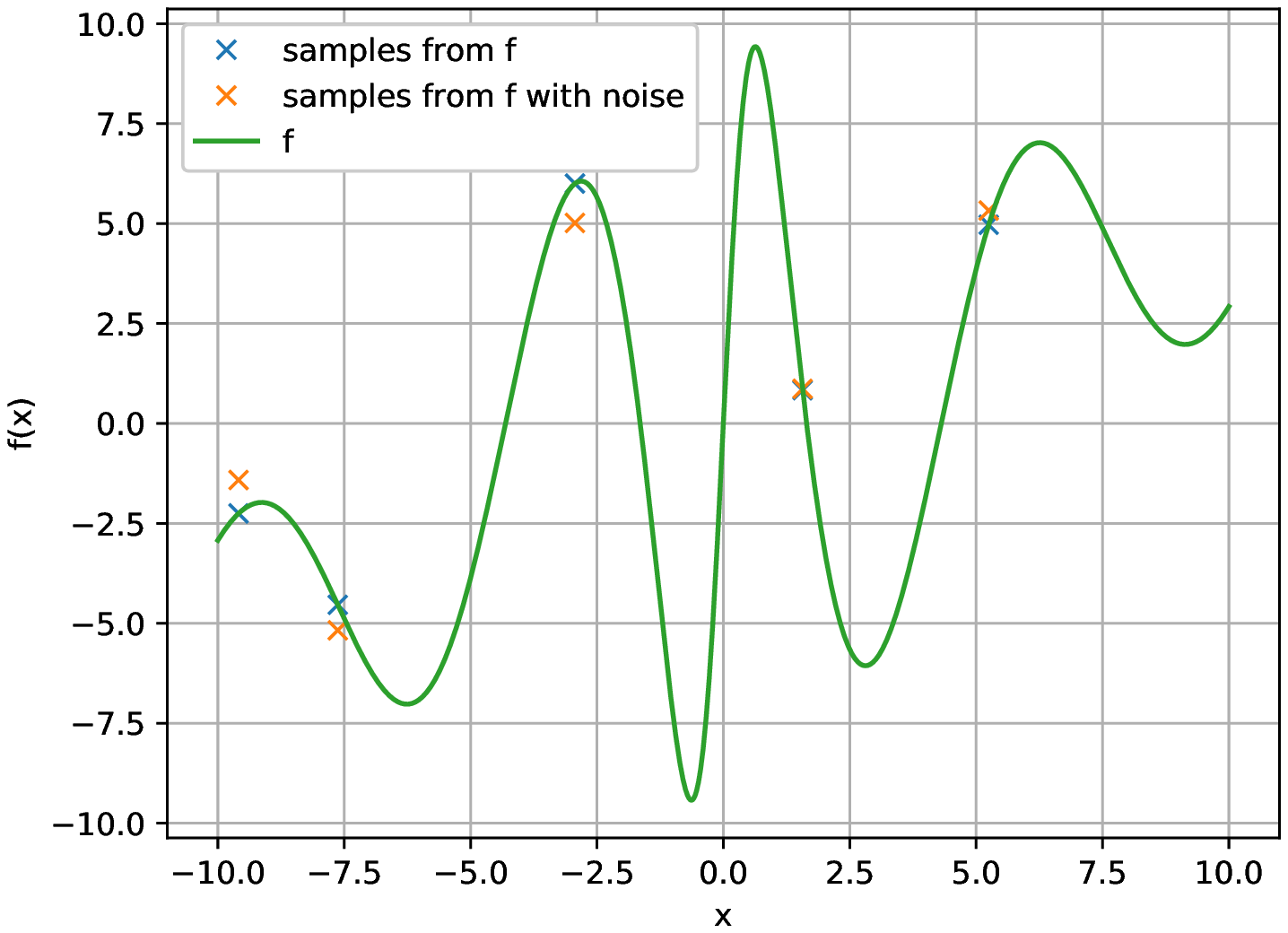}\label{fig:ex_target}}\\
	\subfloat[The classic GP mean with two standard deviations. Parameters are optimised on the full history of observations.]{\includegraphics[width=0.45\textwidth]{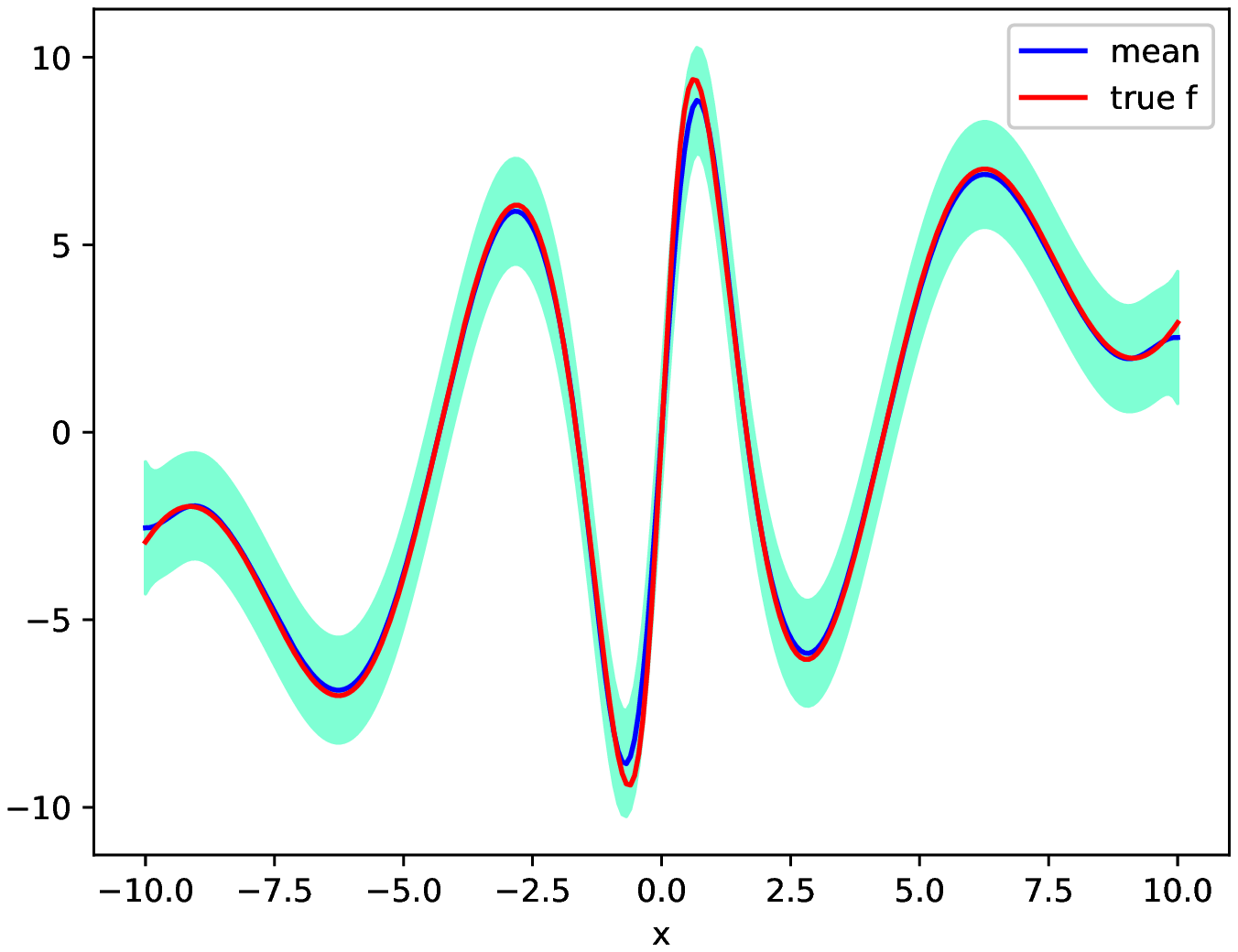}\label{fig:normal_gp}}
\caption{Target function and classic GP approximation}
\label{fig:ex_function}
\end{figure}

In this section the performance of the proposed algorithms is evaluated on both synthetic and real data. Three versions of the EnKF for online GP parameters estimation are assessed: Dual GP-EnKF (Section~\ref{sec:dual_dual_enkf_gp}), Liu-West Dual GP-EnKF (Section~\ref{sec:lw_enkf_gp}), and Joint EnKF (Section~\ref{sec:joint_enkf_gp}).  The developed algorithms are compared with the classic GP regression in terms of both computational time and predictive accuracy. At every iteration $t$ the classic GP regression is applied on all historical data.

\begin{figure}[t]
\centering
	\subfloat[Estimated mean, two standard deviation interval of the GP and target function]{\includegraphics[width=0.45\textwidth]{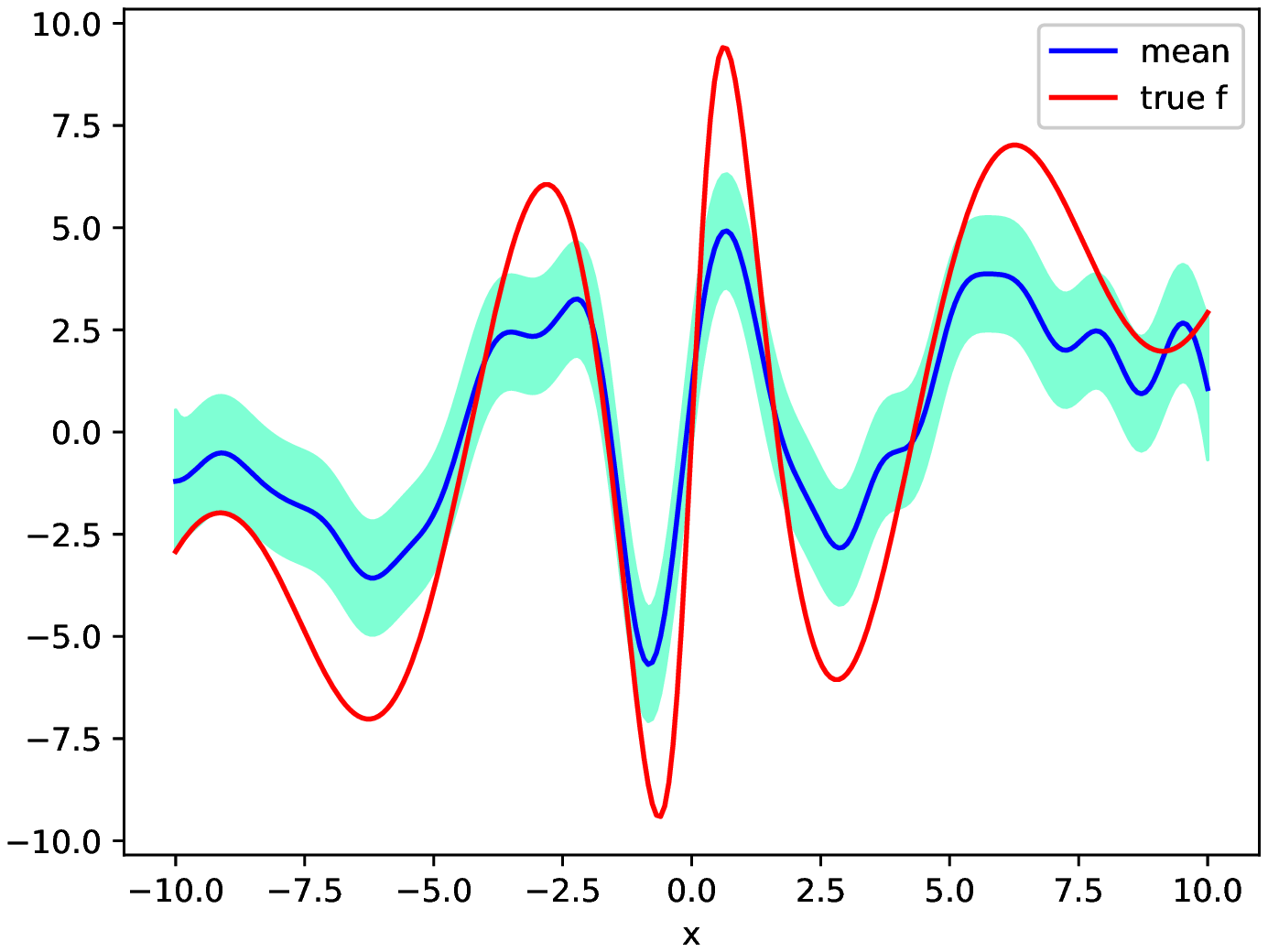}}\\
	\subfloat[Final distributions for the estimated GP parameters]{\includegraphics[width=0.45\textwidth]{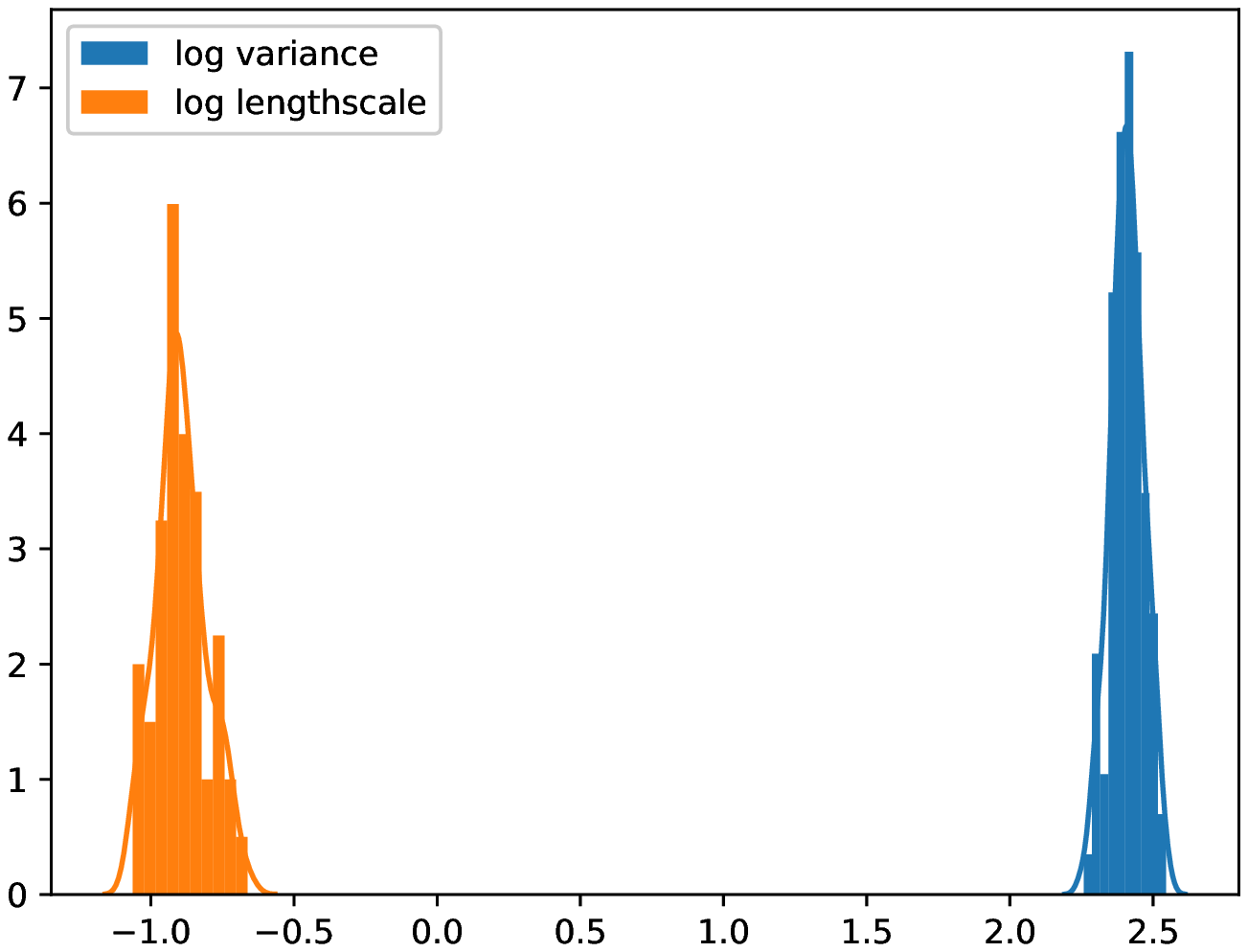}}
	\caption{Performance of Joint GP-EnKF on the synthetic data}
	\label{fig:joint_gp_enkf}
\end{figure}

For quantitative evaluation of the predictive accuracy, the normalised mean squared error of predictions (NMSE) is used on held-out test data $[\mathbf{x}^\text{test}, \mathbf{y}^\text{test}]$:
\begin{equation}
\text{NMSE} = \dfrac{1}{M} \sum_{m=1}^M \frac{\sqrt{(y^\text{test}_m - f^*(\mathbf{x}^\text{test}_m))^2}}{|y^\text{test}_m|},
\end{equation}
where $y^\text{test}_m$ is the observed value of the function at the test data point $\mathbf{x}^\text{test}_m$, $f^*(\mathbf{x}^\text{test}_m)$ is the predicted function value at the test data point.

\subsection{Synthetic data}
The algorithms are firstly evaluated on the synthetic one-dimensional data. The target function for the synthetic data is
\begin{equation}
	f(x) = \frac{x}{2} + \frac{25x}{1+x^2}\cos(x)
\end{equation}
The observations are generated on the domain $[-10, 10]$ and are corrupted by the Gaussian noise with the variance $\sigma^2_y= 0.01$. The example of generated observations is presented in~\figurename~\ref{fig:ex_target}.

All three versions of GP-EnKF use the size of the grid $K = 51$, and ensembles contain $N=100$ members. The covariance function is squared-exponential~\cite{rasmussen2006gaussian}, it has two hyperparameters: $\boldsymbol\theta~=~[\theta_\text{variance},\,\theta_\text{lengthscale}]$ and estimates covariance between two points $x_1$ and $x_2$ as
\begin{equation}
K(x_1, x_2) = \theta_\text{variance} \exp\left\{-\frac{||x_1 - x_2||^2_2}{\theta_\text{lengthscale}^2}\right\}
\end{equation}

At every iteration, $S = 5$ samples are fed into the algorithms, the total number of iterations is $T = 200$.

\figurename~\ref{fig:normal_gp} shows the function estimate given by the classic GP regression. Since the total number of observations is sufficiently large, the classic GP is enable to reconstruct ideal predictions of the function.

\begin{figure}[t]
\centering
	\subfloat[Estimated mean, two standard deviation interval of the GP prediction and target function]{\includegraphics[width=0.45\textwidth]{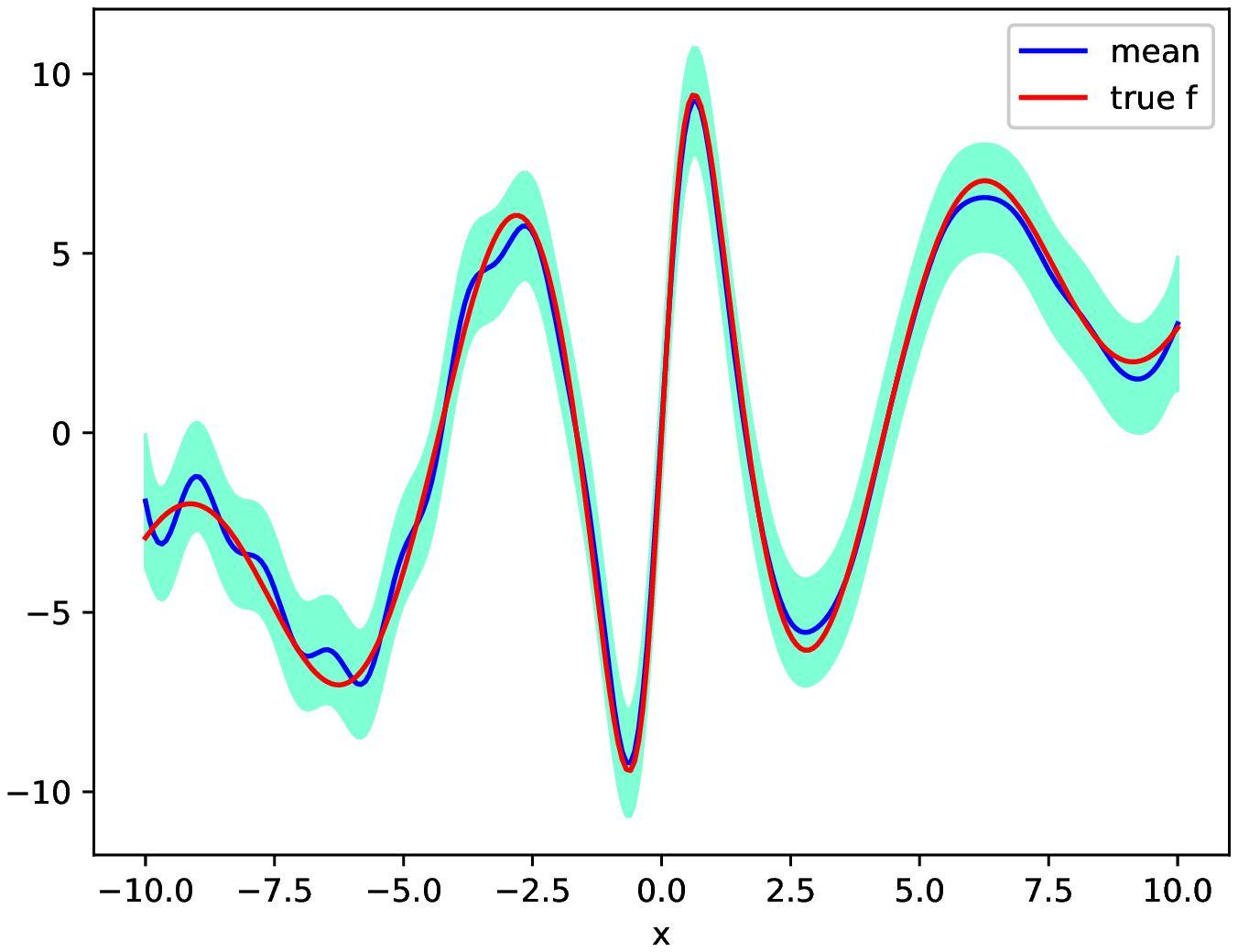}}\\
	\subfloat[Final distributions for the estimated GP parameters]{\includegraphics[width=0.45\textwidth]{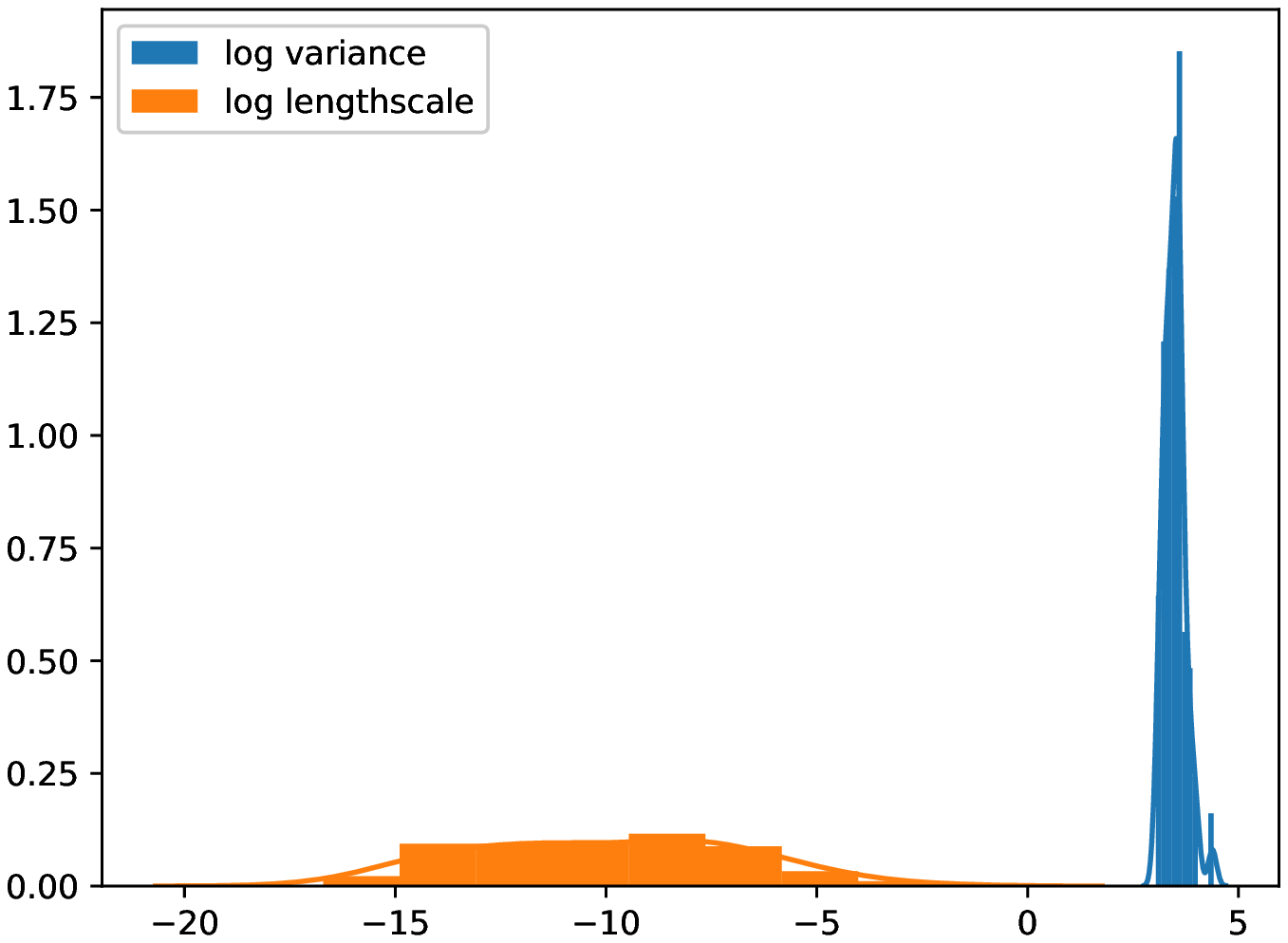}}
	\caption{Performance of Dual GP-EnKF on the synthetic data}
	\label{fig:dual_gp_enkf}
\end{figure}

\begin{figure}[t]
\centering
	\subfloat[Estimated mean, two standard deviation interval of the GP and target function]{\includegraphics[width=0.45\textwidth]{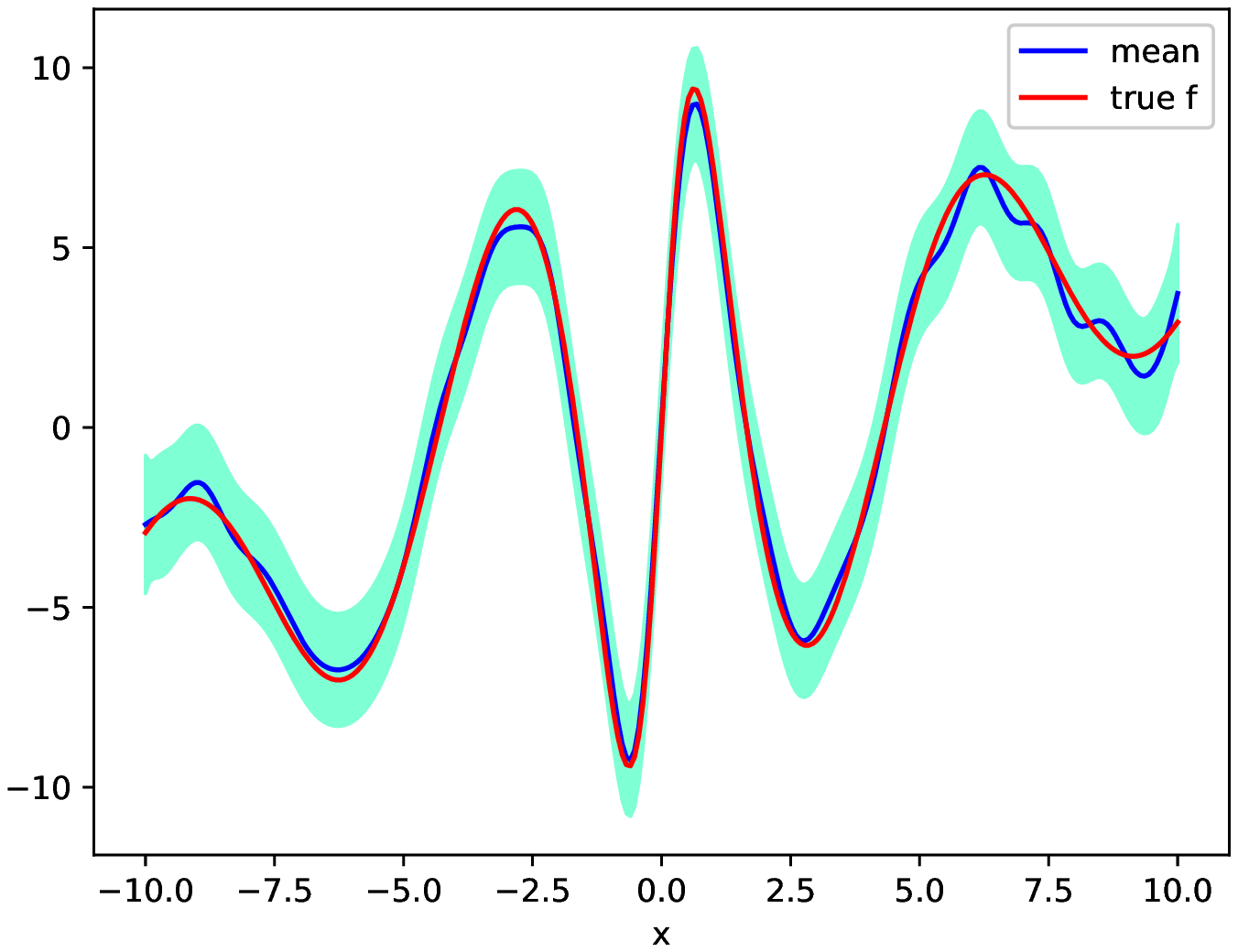}}\\
	\subfloat[Final distributions for the estimated GP parameters]{\includegraphics[width=0.45\textwidth]{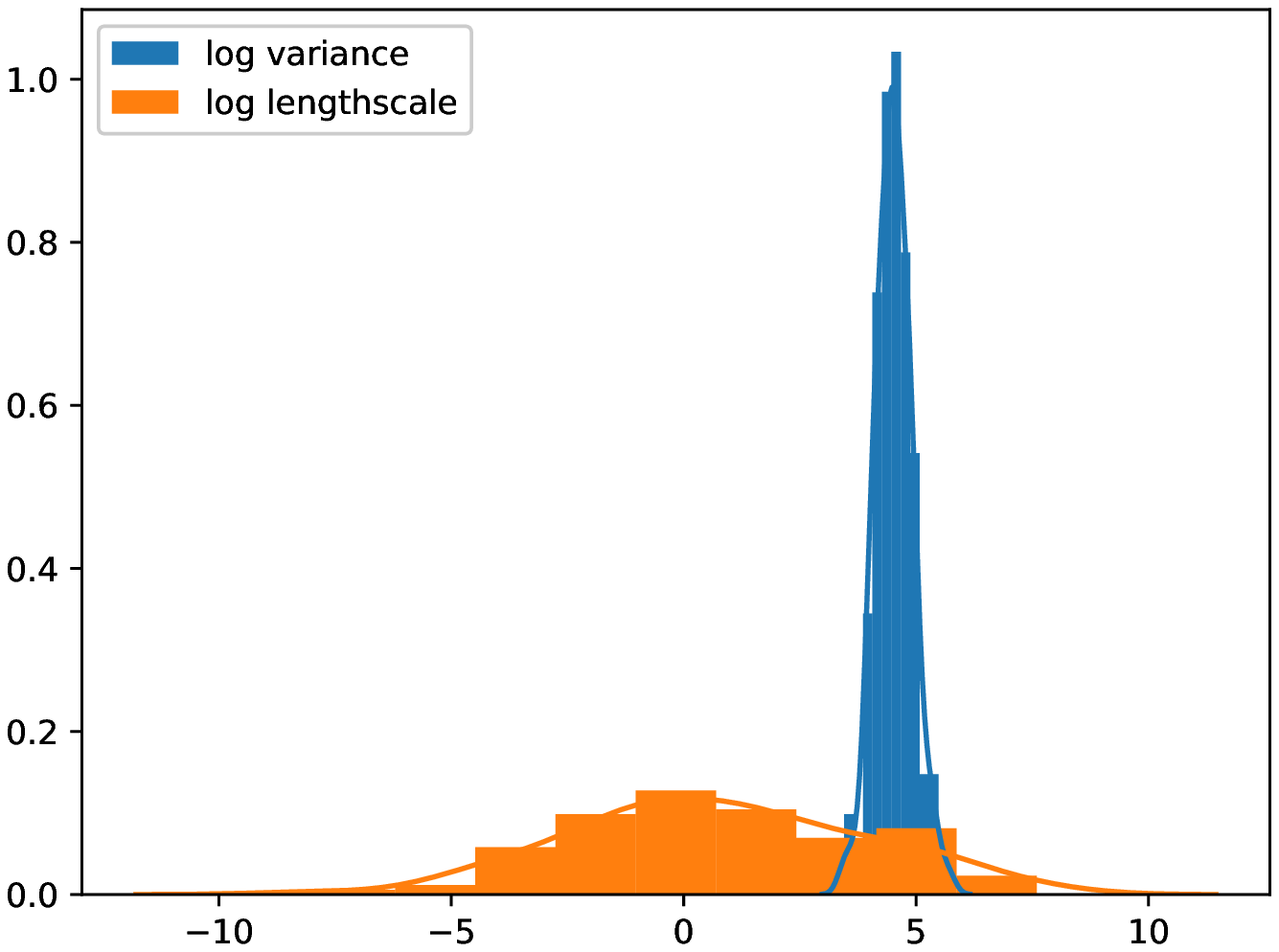}}
	\caption{Performance of the Liu-West Dual GP-EnKF on the synthetic data}
	\label{fig:liuwest_gp_enkf}
\end{figure}

The performance of the proposed approaches is given in \figurename~\ref{fig:joint_gp_enkf}--\ref{fig:liuwest_gp_enkf}. Joint GP-EnKF (\figurename~\ref{fig:joint_gp_enkf}) correctly estimates peaks of the target function, but it has large predictive errors for most of the observations. Joint GP-EnKF learns the consistent ensemble estimates of the hyperparameters, i.e. their variance is not large.

Dual GP-EnKF~(\figurename~\ref{fig:dual_gp_enkf}) provides predictions that are more accurate than predictions made by Joint GP-EnKF, but still there are several locations where the shape of the target function differs from the predicted mean. The ensemble of Dual GP-EnKF has low variance for the logarithm of the lengthscale hyperparameter of the covariance function and high variance for the estimates of the variance hyperparameter.

Liu-West Dual GP-EnKF (\figurename~\ref{fig:liuwest_gp_enkf}) is applied with discount factor $\delta_{lw} = 0.95$. The algorithm makes predictions that are closer to the true values of the target function than other algorithms. The ensemble of Liu-West Dual GP-EnKF gives better estimations of hyperparameters than both Dual and Joint GP-EnKFs.

\begin{figure}[t]
\centering
	\includegraphics[width=0.45\textwidth]{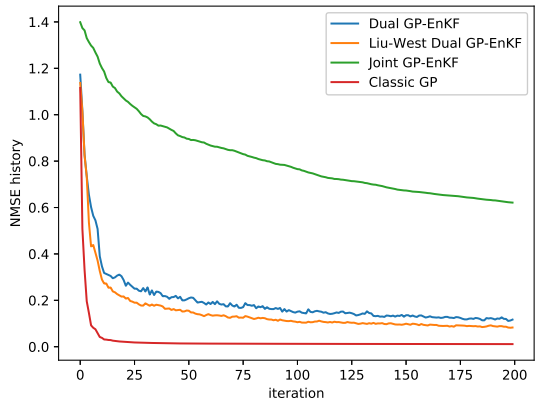}
  \caption{History of NMSE}
	\label{fig:convergence_quality_history}
\end{figure}

The procedure is repeated for $10$ Monte Carlo runs with different random seeds. The results are presented as average among these $10$ Monte Carlo runs. In the \figurename\ref{fig:convergence_quality_history} the history of NMSE is given over time. Its final values together with the computational time are presented in Table~\ref{tab:synthetic_performance}. While Joint GP-EnKF is the fastest method, NMSE of both Dual GP-EnKF methods is lower, and Liu-West Dual GP-EnKF provides the best results. The classic GP approach provides the lowest NMSE, however, it has the computational time more than $10$ times higher than of the slowest of the proposed approaches.

 \begin{table}[b]
  \caption{Performance on the synthetic data at $T=200$}
  \label{tab:synthetic_performance}
  \centering
  \begin{tabular}{llll}
    \toprule
    Method & NMSE &Time (s)\\
    \midrule
    Joint GP-EnKF & $0.64$ &  $\mathbf{7.23}$\\
    Dual GP-EnKF & $0.48$ &   $13.68$\\
    Liu-West Dual GP-EnKF & $\mathbf{0.19}$ &  $15.60$\\
    \midrule
    classic GP & $0.02$ &  $186.20$\\
    \bottomrule
  \end{tabular}
\end{table}

\subsection{House Prices}
The Dual GP-EnKF approach is further evaluated on the real data. In this example the HM Land Registry Price Paid Data\footnote{https://data.gov.uk/dataset/land-registry-monthly-price-paid-data/} is considered. The subset of all flats and maisonettes sold in 2017 is selected and the parameter estimation is performed to predict mean prices corresponding to the locations of properties. Longitude and latitude values for every location have been calculated based on the postcode. Therefore, in this experiment, every single input $\mathbf{x}$ is two-dimensional.

A total of $T = 20$ iterations have been performed with two-dimensional grid of size $K=25 \times 25=625$. At every iteration, $S=100$ samples of the logarithms of standardised prices are used to update parameters and mean in the grid points. The ensemble consists of $N=200$ members. The covariance function is stationary squared-exponential.

\begin{figure}
	\centering
	\subfloat[Mean prices after the first batch of data is presented to the algorithm]{\includegraphics[width=0.45\textwidth]{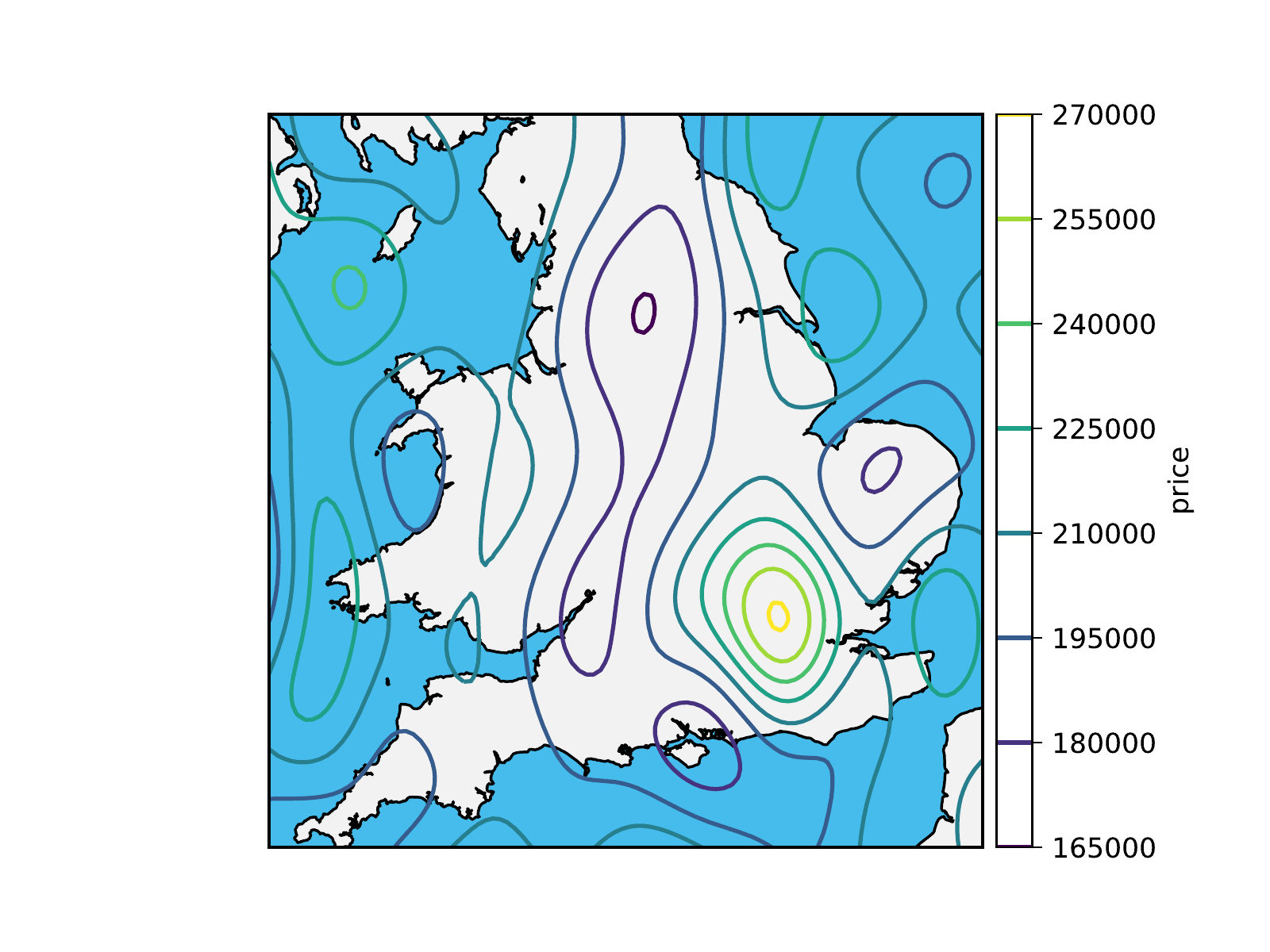}}\\
	\subfloat[Mean prices after the 20th batch of data is presented to the algorithm]{\includegraphics[width=0.45\textwidth]{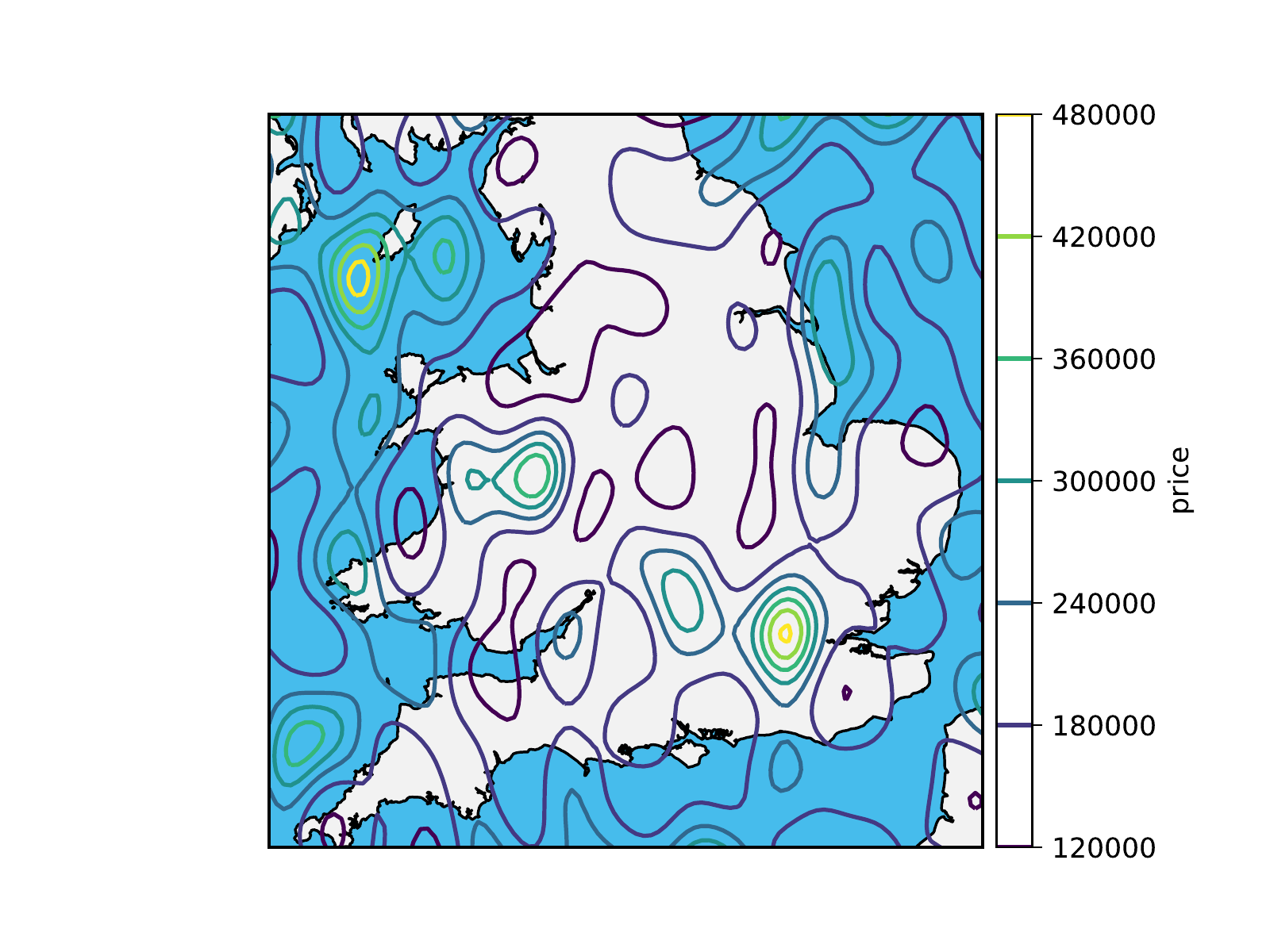}}

    	\caption{Mean estimates of the prices with Dual GP-EnKF}
	\label{fig:house_prices_example}
\end{figure}

\figurename\ref{fig:house_prices_example} demonstrates the results after the first and final iterations. It is clear that the prices have converged close to real values, identifying such areas as London and Oxford as places with higher prices. Though there are spikes of the mean in the sea, the corresponding covariance values that describe uncertainty of predictions in these points are high. Note that the used squared-exponential covariance function is one of the simplest covariance functions in terms of complexity of modelling dependencies of function values at different data points. The stationary squared-exponential covariance function does not depend on locations. Therefore, the results can potentially be further improved if the squared-exponential covariance function is considered together with non-stationary covariance functions to obtain more precise estimates for covariance difference between sea and land locations.

\section{Conclusions}
\label{sec:conclusion}
The paper proposes two ensemble Kalman filters for online Gaussian process regression and learning. The mean and hyperparameters of the GP are interpreted as the state and parameters of the ensemble Kalman filter, respectively. The ensemble Kalman filter updates are utilised to recursively improve estimates of both state and parameters. Two versions of the ensemble updates are proposed: Joint GP-EnKF where the update step of the EnKF is applied for the augmented vector-parameter state and Dual GP-EnKF where the update step is split to first update the parameters and then based on new estimates of the parameters the state is updated. For the Dual EnKF the Liu-West filter~\cite{liu2001combined} updates are additionally developed for further improvement of the estimates.

The proposed ensemble Kalman filter approach for the GP has a linear computational complexity with respect to the number of sequential observations, it depends mainly on the dimensionality of the observations at each timestamp and internal parameters of the filter. For the large volume of data acquired sequentially, it can significantly reduce the computational time in comparison to the usual GP regression that scales cubically with respect to the number of observations. Starting from sufficient number of observations, cubic complexity makes the usual GP not applicable for this large-scale data. The proposed ensemble Kalman filter can be used with any number of sequential observations given that at each timestamp the dimensionality of observations is feasible.

The experiments both on synthetic and real data show that the proposed ensemble Kalman filter approaches for the Gaussian process estimation provide satisfactory predictive accuracy using significantly less computational time in comparison to the GP regression without online updates. Among the proposed approaches the Liu-West Dual GP-EnKF filter demonstrates the best results in terms of the predictive accuracy slightly underperforming the Joint EnKF in terms of the computational time.

\subsubsection*{Acknowledgments}
The authors would like to thank the support from the EC Seventh Framework Programme [FP7 2013-2017] TRAcking in compleX sensor systems (TRAX) Grant agreement no.: 607400.

\bibliographystyle{IEEEtran}
\bibliography{IEEEabrv,bibliography}

\begin{thebibliography}{10}
\providecommand{\url}[1]{#1}
\csname url@samestyle\endcsname
\providecommand{\newblock}{\relax}
\providecommand{\bibinfo}[2]{#2}
\providecommand{\BIBentrySTDinterwordspacing}{\spaceskip=0pt\relax}
\providecommand{\BIBentryALTinterwordstretchfactor}{4}
\providecommand{\BIBentryALTinterwordspacing}{\spaceskip=\fontdimen2\font plus
\BIBentryALTinterwordstretchfactor\fontdimen3\font minus
  \fontdimen4\font\relax}
\providecommand{\BIBforeignlanguage}[2]{{%
\expandafter\ifx\csname l@#1\endcsname\relax
\typeout{** WARNING: IEEEtran.bst: No hyphenation pattern has been}%
\typeout{** loaded for the language `#1'. Using the pattern for}%
\typeout{** the default language instead.}%
\else
\language=\csname l@#1\endcsname
\fi
#2}}
\providecommand{\BIBdecl}{\relax}
\BIBdecl

\bibitem{rasmussen2006gaussian}
C.~Rasmussen and C.~Williams, \emph{Gaussian Processes for Machine
  Learning}.\hskip 1em plus 0.5em minus 0.4em\relax MIT Press, 2006.

\bibitem{turner2011demodulation}
R.~E. Turner and M.~Sahani, ``Demodulation as probabilistic inference,''
  \emph{IEEE Transactions on Audio, Speech, and Language Processing}, vol.~19,
  no.~8, pp. 2398--2411, 2011.

\bibitem{perez2013gaussian}
F.~P{\'e}rez-Cruz, S.~Van~Vaerenbergh, J.~J. Murillo-Fuentes,
  M.~L{\'a}zaro-Gredilla, and I.~Santamaria, ``Gaussian processes for nonlinear
  signal processing: An overview of recent advances,'' \emph{IEEE Signal
  Processing Magazine}, vol.~30, no.~4, pp. 40--50, 2013.

\bibitem{svensson2015marginalizing}
A.~Svensson, J.~Dahlin, and T.~B. Sch{\"o}n, ``Marginalizing {G}aussian process
  hyperparameters using sequential {M}onte {C}arlo,'' in \emph{2015 IEEE 6th
  International Workshop on Computational Advances in Multi-Sensor Adaptive
  Processing (CAMSAP)}.\hskip 1em plus 0.5em minus 0.4em\relax IEEE, 2015, pp.
  477--480.

\bibitem{quinonero2005unifying}
J.~Qui{\~n}onero-Candela and C.~E. Rasmussen, ``A unifying view of sparse
  approximate {G}aussian process regression,'' \emph{Journal of Machine
  Learning Research}, vol.~6, no. Dec, pp. 1939--1959, 2005.

\bibitem{titsias2009variational}
M.~Titsias, ``Variational learning of inducing variables in sparse {G}aussian
  processes,'' in \emph{Artificial Intelligence and Statistics}, 2009, pp.
  567--574.

\bibitem{bui2016unifying}
T.~D. Bui, J.~Yan, and R.~E. Turner, ``A unifying framework for {G}aussian
  process pseudo-point approximations using power expectation propagation,''
  \emph{Journal of Machine Learning Research}, vol.~18, no. 104, pp. 1--72,
  2017.

\bibitem{shen2006fast}
Y.~Shen, M.~Seeger, and A.~Y. Ng, ``Fast {G}aussian process regression using
  {KD}-trees,'' in \emph{Advances in neural information processing systems},
  2006, pp. 1225--1232.

\bibitem{gal2014distributed}
Y.~Gal, M.~Van Der~Wilk, and C.~E. Rasmussen, ``Distributed variational
  inference in sparse {G}aussian process regression and latent variable
  models,'' in \emph{Advances in Neural Information Processing Systems}, 2014,
  pp. 3257--3265.

\bibitem{huber2014recursive}
M.~F. Huber, ``Recursive {G}aussian process: On-line regression and learning,''
  \emph{Pattern Recognition Letters}, vol.~45, pp. 85--91, 2014.

\bibitem{yin2017received}
F.~Yin, Y.~Zhao, F.~Gunnarsson, and F.~Gustafsson, ``Received-signal-strength
  threshold optimization using {G}aussian processes,'' \emph{IEEE Transactions
  on Signal Processing}, vol.~65, no.~8, pp. 2164--2177, 2017.

\bibitem{yin2017distributed}
F.~Yin and F.~Gunnarsson, ``Distributed recursive {G}aussian processes for
  {RSS} map applied to target tracking,'' \emph{IEEE Journal of Selected Topics
  in Signal Processing}, vol.~11, no.~3, pp. 492--503, 2017.

\bibitem{zhao2016gaussian}
Y.~Zhao, F.~Yin, F.~Gunnarsson, F.~Hultkratz, and J.~Fagerlind, ``Gaussian
  processes for flow modeling and prediction of positioned trajectories
  evaluated with sports data,'' in \emph{19th International Conference on
  Information Fusion (FUSION)}.\hskip 1em plus 0.5em minus 0.4em\relax IEEE,
  2016, pp. 1461--1468.

\bibitem{murray2010slice}
I.~Murray and R.~P. Adams, ``Slice sampling covariance hyperparameters of
  latent {G}aussian models,'' in \emph{Advances in Neural Information
  Processing Systems}, 2010, pp. 1732--1740.

\bibitem{osborne2012real}
M.~A. Osborne, S.~J. Roberts, A.~Rogers, and N.~R. Jennings, ``Real-time
  information processing of environmental sensor network data using {b}ayesian
  {G}aussian processes,'' \emph{ACM Transactions on Sensor Networks (TOSN)},
  vol.~9, no.~1, p.~1, 2012.

\bibitem{evensen1994sequential}
G.~Evensen, ``Sequential data assimilation with a nonlinear quasi-geostrophic
  model using {M}onte {C}arlo methods to forecast error statistics,''
  \emph{Journal of Geophysical Research: Oceans}, vol.~99, no.~C5, pp.
  10\,143--10\,162, 1994.

\bibitem{roth2017ensemble}
M.~Roth, G.~Hendeby, C.~Fritsche, and F.~Gustafsson, ``The ensemble {K}alman
  filter: a signal processing perspective,'' \emph{EURASIP Journal on Advances
  in Signal Processing}, vol. 2017, no.~1, p.~56, 2017.

\bibitem{anderson2001ensemble}
J.~L. Anderson, ``An ensemble adjustment {K}alman filter for data
  assimilation,'' \emph{Monthly weather review}, vol. 129, no.~12, pp.
  2884--2903, 2001.

\bibitem{evensen2009ensemble}
G.~Evensen, ``The ensemble {K}alman filter for combined state and parameter
  estimation,'' \emph{IEEE Control Systems}, vol.~29, no.~3, 2009.

\bibitem{wan1997dual}
E.~A. Wan and A.~T. Nelson, ``Dual {K}alman filtering methods for nonlinear
  prediction, smoothing and estimation,'' in \emph{Advances in neural
  information processing systems}, 1997, pp. 793--799.

\bibitem{moradkhani2005dual}
H.~Moradkhani, S.~Sorooshian, H.~V. Gupta, and P.~R. Houser, ``Dual
  state--parameter estimation of hydrological models using ensemble {K}alman
  filter,'' \emph{Advances in water resources}, vol.~28, no.~2, pp. 135--147,
  2005.

\bibitem{mitchell2000adaptive}
H.~L. Mitchell and P.~Houtekamer, ``An adaptive ensemble {K}alman filter,''
  \emph{Monthly Weather Review}, vol. 128, no.~2, pp. 416--433, 2000.

\bibitem{delsole2010state}
T.~DelSole and X.~Yang, ``State and parameter estimation in stochastic
  dynamical models,'' \emph{Physica D: Nonlinear Phenomena}, vol. 239, no.~18,
  pp. 1781--1788, 2010.

\bibitem{stroud2007sequential}
J.~R. Stroud and T.~Bengtsson, ``Sequential state and variance estimation
  within the ensemble {K}alman filter,'' \emph{Monthly Weather Review}, vol.
  135, no.~9, pp. 3194--3208, 2007.

\bibitem{liu2001combined}
J.~Liu and M.~West, ``Combined parameter and state estimation in
  simulation-based filtering,'' in \emph{Sequential Monte Carlo methods in
  practice}.\hskip 1em plus 0.5em minus 0.4em\relax Springer, 2001, pp.
  197--223.

\end{thebibliography}

\end{document}